\RequirePackage{fix-cm}
\documentclass[smallextended]{svjour3}       
\smartqed  
\usepackage{natbib}
\usepackage{graphicx}
\usepackage[fleqn]{amsmath}
\usepackage{color}
\newcommand{\etal}{\textit{et al.}}
\newcommand{\eg}{\textit{e.g.}}
\newcommand{\ie}{\textit{i.e.}}

\newcommand{\argmax}{\mathop{\rm arg~max}\limits}

 \journalname{myjournal}
\begin{document}

\title{Zero-resource Machine Translation by Multimodal Encoder-decoder Network with Multimedia Pivot}

\titlerunning{Zero-resource Machine Translation with Multimedia Pivot}        

\author{Hideki Nakayama         \and
        Noriki Nishida 
}


\institute{H. Nakayama \at
              Graduate School of Information Science and Technology, The University of Tokyo \\
              1-1-1, Yayoi, Bunkyo-ku, Tokyo, Japan \\
              Tel.: +81-3-58418986\\
              FAX: +81-3-58418986\\
              \email{nakayama@ci.i.u-tokyo.ac.jp}           
           \and
           N. Nishida \at
              Graduate School of Information Science and Technology, The University of Tokyo \\
              1-1-1, Yayoi, Bunkyo-ku, Tokyo, Japan \\
              \email{nishida@nlab.ci.i.u-tokyo.ac.jp}           
}

\date{Received: date / Accepted: date}

\maketitle

\begin{abstract}
We propose an approach to build a neural machine translation system with no supervised resources (\ie, no parallel corpora) using multimodal embedded representation over texts and images.
Based on the assumption that text documents are often likely to be described with other multimedia information (\eg, images) somewhat related to the content, we try to indirectly estimate the relevance between two languages.
Using multimedia as the "pivot", we project all modalities into one common hidden space where samples belonging to similar semantic concepts should come close to each other, whatever the observed space of each sample is. 
This modality-agnostic representation is the key to bridging the gap between different modalities. 
Putting a decoder on top of it, our network can flexibly draw the outputs from any input modality. 
Notably, in the testing phase, we need only source language texts as the input for translation. 

In experiments, we tested our method on two benchmarks to show that it can achieve reasonable translation performance.
We compared and investigated several possible implementations and found that an end-to-end model that simultaneously optimized both rank loss in multimodal encoders and cross-entropy loss in decoders performed the best.
\end{abstract}

\section{Introduction}
\label{intro}
Machine translation (MT) has been one of the most important challenges in natural language processing. 
Irrespective of traditional statistical machine translation (SMT) \citep{Koehn2009} or modern neural machine translation (NMT) \citep{Sutskever2014}, 
methods and data have always been mutually indispensable to each other.
Indeed, the success of corpus-based MT is mainly dependent on the quality and scale of available parallel corpora to train MT systems.
Recent state-of-the-art NMT systems have shown that translation can be surprisingly improved with sufficiently large-scale data and high computational power \citep{Shen2016}.

On the other hand, how to prepare such corpora has remained a big problem.
In some specific domains such as Web news, patents, and Wikipedia, relatively high-quality multilingual translations are made available by content holders or volunteer workers, which have been utilized by researchers for decades \citep{Koehn2005,Taeger2011}.
However, in more general cases, it is not always possible to collect a sufficient amount of parallel data because most generic Web documents are monolingual.
The human cost for preparing manual translation is quite high, and it is particularly prohibitive for minor language pairs where resources are severely limited.

To tackle the situation where no or only a few parallel corpora are available, a branch of MT called pivot-based machine translation has been developed. 
The idea of the pivot-based approach is to indirectly learn the alignment of the source and target languages with the help of a third modality (\eg, texts in another language). 
Although previous studies along this line have been mainly based on the third language,  
in this work, we propose a novel and more general framework to utilize arbitrary multimedia content (\eg, images) as the pivot.
Nowadays, we can easily find abundant monolingual text documents with rich multimedia content as the side information,
\eg, text with photos or videos posted to social networking sites and blogs. 
These visual media are expected to be more or less correlated to the counterpart texts following the objective of a document. 
Considering that we can generally understand the content of images taken in other countries regardless of our own language, visual information can be a universal representation to ground different languages.

Moreover, in recent years, performance of visual recognition has been dramatically improved owing to the huge success of deep learning,
where it is now considered to be on a human level for generic image recognition \citep{Krizhevsky2012}. 
We expect that these state-of-the-art visual recognition techniques are now mature enough to accurately extract language-agnostic semantics of images to help improve natural language processing (NLP) tasks.
If multimedia pivot-based machine translation is established, we could possibly utilize abundant monolingual multimedia documents naturally provided by Web users to build high-performance and open-domain MT systems.

Our contributions in this study are as follows:
\begin{enumerate}
\item To the best of our knowledge, we are the first to propose a zero-resource (\ie, no direct parallel corpus) machine translation method that utilizes multimedia as the pivot. Importantly, pivot images are required only in the training phase.
\item To realize this, we propose a neural network based method combining multimodal (cross-modal) representation learning and encoder-decoder models. 
We note that our model can align source encoder and target decoder without source-to-target path during training which is often utilized by pseudo corpus based methods. 
Moreover, our idea is agnostic to implementations of encoder and decoder networks.
\item We categorized several possible approaches in model topology and learning strategies and extensively investigated their performance.  
\end{enumerate}

\vspace{-1mm}
\section{Related Work}
\vspace{-1mm}
\subsection{Resource Problem in Cross-lingual Learning}
\vspace{-1mm}
Dealing with limitations in the number of good-quality parallel or comparable corpora has been one of the most important issues in cross-lingual learning.
One straightforward approach is to 
automatically mine parallel corpora, typically from noisy Web repositories. 
Some methods exploited a bootstrap approach starting from base translation systems \citep{Uszkoreit2010}, whereas others utilized external meta information such as links to the same URL to coupling bilingual texts~\citep{Riesa2012}.
Among them, images have also been exploited as a key for cross-lingual document matching in relatively early works. However, these methods simply rely on OCR reading or near-duplicate (copy) detection of images \citep{Oard1999}, 
and thus they cannot identify similarities in semantics, which is a fundamental limitation as compared to our work. 

Another line of work has been to train MT system from non-parallel data with the help of another modality for indirect knowledge transfer, which is called the pivot-based machine translation.
Most recent works have focused on existing popular language to use as the pivot \citep{Wu2007,Wu2009,Firat2016}.
While creating direct parallel corpora in minor language pairs is practically very difficult, major languages (\eg, English) are relatively often coupled to each language.
Source-to-target translation can be realized by first translating the source language into the pivot language and then translating it into the target language.
Nonetheless, this method still assumes that source-pivot and pivot-target parallel corpora are available, which would require the effort of human experts if the languages are minor ones.
Moreover, it is difficult to use images as the pivot in this approach because explicitly decoding an image from text is not a well-established technique.
Therefore, image-based pivots have mainly been used in relatively easier tasks such as bilingual lexicon learning, where image similarity is used as the criteria to estimate relevance between tag words attached to images \citep{Bergsma2011,Kiela2015,Vuli2016}. 

\vspace{-1mm}
\subsection{Computer Vision for Machine Translation}
\vspace{-1mm}
Grounding a natural language to real-world representations has always been an important topic in NLP, for which computer vision would be the first natural choice \citep{Silberer2014}. 
After a huge breakthrough in the use of convolutional neural networks (CNNs) \citep{Krizhevsky2012}, visual recognition has been significantly advanced in terms of both accuracy and flexibility, 
enabling the development of many brand-new technologies.
Amongst them, image captioning, which automatically annotates a description for an input image with natural language, has become one of the hottest topics in recent years \citep{Vinyals2015,Johnson2016}.
Because image captioning is essentially interpreted as "translation" from an image to sentence, it has drawn more and more attention in the NLP community as well.

Recently, a new research field called multimodal machine translation was proposed \citep{Elliott2015,Hitschler2016}, which became a subtask in WMT 2016\footnote{http://www.statmt.org/wmt16/multimodal-task.html}\citep{Specia2016}.
The aim of this task is to use images in addition to source languages as inputs to improve the translation performance, hopefully relaxing ambiguity in alignment that cannot be solved by texts only.  
The feasibility of this approach has been demonstrated by some methods, such as visual-based reranking of SMT results \citep{Hitschler2016}.
However, this task assumes that images are available as a part of a query in the testing phase, and thus the objective and setup are entirely different from ours.

\vspace{-1mm}
\subsection{Multimodal Embedding}
\vspace{-1mm}
To use non-language multimedia as the pivot for MT, we need a more flexible mechanism to semantically align different types of data.
The key idea here is to derive one common representation shared by all modalities.
In other words, whatever the modality is, observed data belonging to the same implicit concept should be mapped into roughly the same point in the embedding space. 
The most classical and standard method for multimodal learning is probably linear canonical correlation analysis (CCA) \citep{Hotelling1936},
which has been successfully used in image-language collaborations such as semantic image retrieval and annotation \citep{Hardoon2004}, 
as well as cross-lingual information retrieval \citep{Udupa2010} \citep{Funaki2015}. 
In more recent methods based on deep neural networks, pairwise ranking loss has been shown to significantly improve multimodal embedding \citep{Frome2013} owing to its natural capability of learning discriminative nearest-neighbor metrics and stability in gradient-based learning. 
It was successfully used for image captioning within the framework of the deep encoder-decoder model \citep{Kiros2014a}.

In this work, we simultaneously optimize source-pivot~(image) and pivot-target losses with a shared pivot encoder to implicitly align two languages in the multimodal space, 
which is the core of our zero-shot learning.
We further put a target sequence decoder on top of the multimodal representation to compose an end-to-end encoder-decoder model.
From the theoretical viewpoint, our work is in the line of some recently proposed methods in the form of multi-stream encoder-decoder model. 
We look into these models and ours in detail and describe our contribution in Section \ref{sec:diff}.

\begin{figure*}
\begin{center}
\includegraphics[width=1.01\columnwidth]{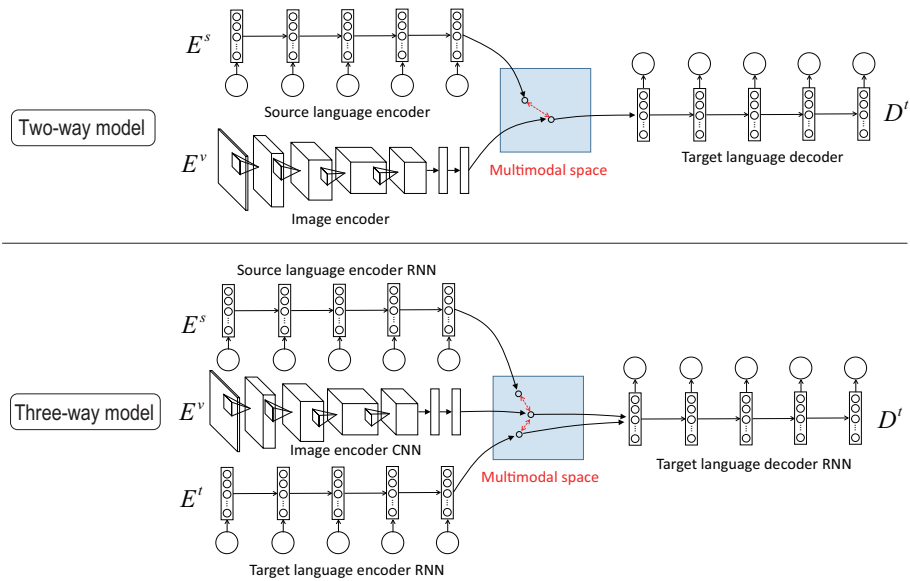}
\caption{Our models for neural translation based on pivot images (Top: two-way model. Bottom: three-way model). In the training phase, language encoders $E^s$ and $E^t$ are forced to have high correlations with the image encoder $E^v$ in the multimodal space, on which the decoder of the target language $D^t$ is trained. In the testing phase, translation can be realized by simply feedforwarding through $E^s$ and $D^t$.} 
\label{fig:models}
\end{center}
\end{figure*}

\vspace{-1mm}
\section{Our Approach}
\vspace{-1mm}
\subsection{Overview}
\vspace{-1mm}
Our goal is to build a translation model from a source language $s$ to a target language $t$ by utilizing the side information (images) as the pivot.
Below, we call a pair of a text description $d$ and its counterpart image $i$ a "document."
For training the system, suppose that we have $N^s$ monolingual documents in the source language, $\mathcal{T}^s=\{d^s_k,i^s_k\}^{N^s}_{k=1}$.
Similarly, we also have $N^t$ documents in the target language, $\mathcal{T}^t=\{d^t_k,i^t_k\}^{N^t}_{k=1}$.
Importantly, $\mathcal{T}^s$ and $\mathcal{T}^t$ do not overlap; they do not share the same images at all.
While $d^s$ and $d^t$ obviously appear in different spaces, $i^s$ and $i^t$ share a common visual space and can be handled by the same encoder.
We let $E^s(d^s)$, $E^t(d^t)$, and $E^v(i)$ denote non-linear encoders (\ie, feature extractors) for source language descriptions, target language descriptions, and images, respectively.

Our model can be divided into roughly two important components. 
The first component is the multimodal representation learning, in which the parameters of the encoders, $E^s(d^s)$, $E^t(d^t)$, and $E^v(i)$, are optimized so that they are mapped into the same semantic space, which we call "the multimodal space." 
If such a good multimodal space is obtained, instances of all modalities should have roughly the same vector representation as long as they are tied together with similar semantic concepts.
The second component is to build a target language decoder, $D^t$, on top of the multimodal space so that the final translation can be realized by $D^t\left(E^s(d^s)\right)$.
It should be emphasized that we only need texts for input during the testing phase, similar to standard machine translation.

Figures \ref{fig:models} illustrates our approach.
There are several options in the model topology and training strategies that are thoroughly compared in the experiments.
We describe the details in the following sections.

\vspace{-1mm}
\subsection{Model Topologies}
\vspace{-1mm}
We use the pair-wise rank loss proposed in \citep{Frome2013} for training encoders to map them to one common multimodal space.  
For the \textit{two-way model}, we take the source-image loss as follows (Fig.~\ref{fig:models}: Top):
\begin{equation}
J_{2w}^E(\mathcal{T}^s) = \sum_{i^s}\sum_{ng} \mbox{max}\{0,\alpha-s\left(E^v(i^s),E^s(d^s)\right)+s\left(E^v(i^s),E^s(d^s_{ng})\right)\},  \label{eqn:im_loss}
\end{equation}
where $\alpha$ is the hyperparameter of margin and the similarity score function, $s()$, measures the dot product. Note that the outputs of each encoder are unit normalized and thus it is equal to cosine similarity. 
 $d^s_{ng}$ denotes negative (not coupled) descriptions for $i^s$ sampled from the same mini-batch. 

For training the decoder, images $i^t$ in $\mathcal{T}^t$ are feedforwarded and used as the inputs to measure decoder loss against $d^t$. We take the standard cross-entropy loss.
\begin{equation}
J^D_{im}(\mathcal{T}^t)= -\sum_{d^t} \frac{1}{|d^t|}\sum_{k=1}^{|d^t|}\mbox{log}P\left(w_k|D^t(E^v(i^t))\right),\end{equation}
where $P(w_k)$ is the probability that the model outputs the ground truth word at step $k$.
Our two-way model is closely related to the image-captioning model proposed by \citep{Kiros2014a} except that we apply different languages to the encoder and decoder parts.
This is viewed as an end-to-end fusion of multimodal embedding and image-captioning models.

In the \textit{three-way model}, we further incorporate rank loss on $\mathcal{T}^t$ in addition to $\mathcal{T}^s$ (Fig.~\ref{fig:models}: Bottom) for training the encoders. 
The encoder loss for the three-way model is defined as follows:
\begin{equation}
\begin{split}
& J_{3w}^E(\mathcal{T}^s,\mathcal{T}^t) = \sum_{i^s}\sum_{ng} \mbox{max}\{0,\alpha-s\left(E^v(i^s),E^s(d^s)\right)+s\left(E^v(i^s),E^s(d^s_{ng})\right)\} \\
& \quad\quad + \sum_{i^t}\sum_{ng} \mbox{max}\{0,\alpha-s\left(E^v(i^t),E^t(d^t)\right)+s\left(E^v(i^t),E^t(d^t_{ng})\right)\}.
\end{split}
\end{equation}

The advantages of the three-way model over the two-way model are many.
First, while image-target alignment is ignored in the two-way model, images implicitly bind source and target languages by jointly enforcing high correlations between them in the three-way model.  
Thus, multimodal representation itself is expected to be improved for bridging the gap between two languages.
Moreover, simultaneously optimizing two constraints would have a positive regularization effect in a manner similar to that of so-called multi-task learning.
Second, unlike in the two-way model, the three-way model can utilize both images and descriptions in $\mathcal{T}^t$ for training decoders of the target language because now they are mapped into a common representation. 
This is interpreted as a sort of data augmentation and is expected to further improve robustness. 
The loss for reconstructing target descriptions is as follows: 
\begin{equation}
J^D_{de}(\mathcal{T}^t)= -\sum_{d^t} \frac{1}{|d^t|}\sum_{k=1}^{|d^t|}\mbox{log}P\left(w_k|D^t(E^t(d^t))\right). \label{eqn:de_loss}
\end{equation}

We can use either $J^D_{im}$, $J^D_{de}$, or both for training the decoder in the three-way approach. 
Now, the model can be viewed as a fusion of multimodal embedding, image captioning, and autoencoder of target languages. 

\vspace{-1mm}
\subsection{Training Strategy}
\vspace{-1mm}
We investigate two strategies for training the whole model.
The first strategy is the \textit{two-step approach}, in which we first optimize the encoder loss, $J^E$. Then we fix the parameters for all encoders and start optimizing the decoder with respect to $J^D$. 
The second strategy is the \textit{end-to-end approach}, in which we jointly optimize encoder and decoder losses. Here we use the combined loss,
\begin{equation}
J^{all} = J^D + \lambda J^E,
\end{equation}
where $\lambda$ is a weighting parameter.

\begin{table}[btp]
 \caption{\label{table:splits}Statistics and data splits for each preprocessed dataset.}
 \begin{center}
  \begin{tabular}{|l|c|c|}
   \hline
                       & IAPR-TC12 & Multi30K \\
   \hline\hline
   \multicolumn{3}{|c|}{Dataset Statistics} \\\hline
   En-im-De triplets & 20000 & 31014\\
   Vocabulary size & En: 1293 & En: 2181 \\
                   & De: 1766 & De: 2171 \\
   Avg. length     & En: 27.3 & En: 13.0 \\
   \;\;of descriptions & De: 22.8 & De: 12.4 \\\hline
   \multicolumn{3}{|c|}{Splits for Experiments} \\\hline
   Train: im-En pairs & 9000 & 14500 \\ 
   Train: im-De pairs  & 9000 & 14500 \\ 
   Val: im-En pairs   & 500 & 507 \\ 
   Val: im-De pairs    & 500 & 507 \\ 
   Test: En-De pairs & 1000 & 1000 \\ 
   \hline
  \end{tabular} 
 \end{center}
\end{table}

\vspace{-1mm}
\subsection{Difference from Closely Related Methods}
\vspace{-1mm}
\label{sec:diff}
Although our work is, as far as we know, the first attempt of zero-resource machine translation using multimedia pivot, 
there have been some theoretically close methods that inspired our model.
The topology of our network is similar to recently proposed many-to-one sequence-to-sequence model \citep{Luong2016}.
However, their model is designed for standard multi-task learning and does not have a multimodal embedding layer like ours. 
Therefore, it cannot align a source encoder and a target decoder in zero-shot situations. 
To deal with zero-shot problem, Firat \etal \citep{Firat2016} incorporated some synthetic parallel corpora to explicitly include source-to-target path during training. 
In other words, they approach the zero-shot problem in data-side while we approach in model-side with the help of multimodal embedding technique.

As for the pivot-based multimodal representation learning,
Funaki \etal~\citep{Funaki2015} and Rajendran \etal~\citep{Rajendran2016} used basically the same idea as our multimodal space, implemented with generalized CCA and neural encoders respectively.
A major difference is that there models have no cross-modal decoders because their interest was the multimodal representation (embedding) itself.
We will show that simultaneously optimizing decoders have positive effects not only for decoding but also for the learned representation itself.
Saha \etal \citep{Saha2016} proposed an end-to-end model of multimodal embedding and target decoder, which is almost identical to our two-way model except that their multimodal fusion is based on correlation loss. As we have described above, our three-way model including the target encoder in multimodal learning have many advantages. In fact, it can significantly improve both the multimodal representation and decoding performance compared to two-way model as we show in the experiments.

\vspace{-1mm}
\section{Experiment}
\vspace{-1mm}
\subsection{Data Set}
\vspace{-1mm}
For our study, we used two publicly available multilingual image-description datasets.
The IAPR-TC12 dataset \citep{Grubinger2006} has 20,000 images with their English and German descriptions.
The original descriptions were provided in German, and their English translations were added by professionals.
The recently published Multi30K dataset \citep{Elliott2016} is specifically designed for research of multimodal machine translation. 
It has 31,014 images with English and German descriptions for each image.
This is an extension of Flickr30K \citep{Young2014}, an image-caption dataset in English, for which German translations are provided by \citep{Elliott2016}.
There are two types of bilingual annotations provided in Multi30K dataset. Namely, one for machine translation task and the other for multilingual image captioning task, respectively. We used the former for our experiments and
followed the official training, validation and testing splits.

For preprocessing, all words were converted into lowercase and tokenized using Natural Language Toolkit, and then those appearing less than 5 times in the training splits were replaced by UNK symbol.
Table \ref{table:splits} summarizes the statistics of the datasets and our experimental setup.
We randomly split data into non-overlapping sets for training, validation, and testing.
Unnecessary modalities for each split (\eg, German descriptions for Image-English split) were ignored.
It is notable that we had no direct English-German parallel data, even in the validation sets.

Although these are the current largest multi-lingual image description datasets as far as we know, 
we should say that they are relatively small compared to standard studies on neural machine translation.
We note that our work is in the beginning stage where our focus is to 
show the feasibility of zero-shot translation using multimedia pivot, as well as to investigate how each component in our model affects the relative improvements in performance.

\vspace{-1mm}
\subsection{Experimental Setup}
\vspace{-1mm}

\begin{table}[btp]
 \caption{\label{table:baseline}BLEU (BLEU+1) scores on supervised baselines (sequence to sequence) changing the size of training data.}
 \begin{center}
  \begin{tabular}{|l||c|c|}
   \hline
            Data Size & IAPR-TC12 & Multi30K \\
   \hline\hline
   \multicolumn{3}{|c|}{De $\rightarrow$ En} \\\hline
    9000/14500 & 45.8 (47.2) & 14.8 (25.1) \\
    3000 & 30.6 (32.9) & 8.9 (18.9) \\
    2000 & 27.3 (29.2) & 7.5 (17.9) \\
    1000 & 24.1 (25.6) & 6.0 (16.9) \\
   \hline
   \multicolumn{3}{|c|}{En $\rightarrow$ De} \\\hline
    9000/14500 & 36.4 (38.3) & 15.7 (27.3) \\
    3000 & 27.9 (28.7) & 9.2 (20.3) \\
    2000 & 23.3 (25.6) & 7.2 (19.2) \\
    1000 & 18.4 (21.4) & 5.8 (17.8) \\
   \hline
  \end{tabular} 
 \end{center}
\end{table}

Because the choice of encoders and decoders for each modality is not within the scope of this paper, we used the most standard neural models for each domain. 
For visual encoder $E^v$, we employed the public VGG-19 network \citep{Simonyan2015}, which is one of the most powerful and widely used CNNs pre-trained on the ImageNet dataset \citep{Deng2009}. We used features from the "fc7" layer of VGG-19 and put another two fully connected (FC) layers with 1024 hidden units each, only which are tuned during the training. 
For implementation, we used the pre-computed features for Multi30K provided at the WMT'16 Multimodal Machine Translation task \footnotemark[1].
We extrracted the same features for IAPR-TC12 using Caffe \citep{Jia2014}.
For language encoders and decoders $E^s$, $E^t$, and $D^t$, we used recurrent neural networks~(RNNs) with long short-term memory (LSTM) \citep{LSTM1997}. 
We used 512-dimensional word embedding and 1024-dimensional hidden units.
Note that the dimensions of all encoders should be equal so that they can be coupled in multimodal space.
We used the Adam optimizer \citep{Kingma2015} with mini-batch size 32 for training the network, and we stopped optimization when the validation loss no longer improved. We fixed $\alpha=0.1$ and $\lambda=100$ through our experiments.

To compose a mini-batch for the language encoders $E^s$, $E^t$ and decoders $D^t$, we padded special \textit{NULL} symbols to align the length of sentences in the same input (or output) batch, which is the standard practice in seq-to-seq learning. As for the image encoder $E^v$, because the visual representation of each example is a static 4096-dimensional vector, a mini-batch is a simple 32 $\times$ 4096 matrix having the feature vectors of batch examples in rows in the same order as the corresponding text-side mini-batch. Training data are randomly shuffled in the beginning of each epoch and then fed into mini-batches in order. 

For evaluation, we mainly used the standard corpus-level BLEU metrics \citep{Papineni2002}. 
We used \textit{multi-bleu.pl} script in Moses toolkit to compute BLEU scores. 
We also evaluated the sentence-level BLEU+1 metrics \citep{Lin2004}, which is a modification of BLEU that has smoothing terms for higher-order n-grams, making it possible to evaluate MT performance on short sentences. 
We note a BLEU score in plain text and corresponding BLEU+1 score in a parenthesis. 

Table \ref{table:baseline} summarizes the results on baseline models.
To demonstrate the performance we could obtain with a supervised parallel corpus,
we show the scores on sequence-to-sequence NMT \citep{Sutskever2014} trained on the same RNN architectures
changing the number of randomly sampled parallel data.

\vspace{-3mm}
\subsection{In-depth Study of Multimodal Space}
\vspace{-1mm}
To separately evaluate the effectiveness of the multimodal space, we first focused on simple nearest-neighbor-based translation. Namely, for a query description in the source language, $d^s_q$, we retrieved its nearest-neighbor training sample in $\mathcal{T}^t$ and then simply output its description. This experiment essentially measures the retrieval performance and is appropriate for evaluating the multimodal representation itself.

As a baseline, we implemented a naive method based on TFIDF and CNN visual features.
For a query, we first retrieved the most similar document in $\mathcal{T}^s$ in terms of cosine similarity of TFIDF text features.
Then, for the coupled image of that document, we retrieved the nearest document in $\mathcal{T}^t$ in terms of the L2 distance of CNN features (\ie, VGG-19 fc7 layer) whose caption would be output as the translation result.

In the multimodal space obtained by our two-way model, we can retrieve the nearest image in the target side, $\mathcal{T}^t$, by computing the dot score, which is the criterion we used in the ranking loss (Eq.~\ref{eqn:im_loss}). 
\begin{equation}
\hat{d}^t_{2w} = d_y^t \;\; \mbox{where}\;\; y = \argmax_k\; s\left(E^v(i^t_k),E^s(d^s_q)\right)
\end{equation}

For the three-way model, in addition to image-based retrieval, we can directly retrieve the nearest description in $\mathcal{T}^t$, which is an interesting characteristic of this model.
\begin{equation}
\begin{split}
& \hat{d}^t_{3w} = d_y^t \;\; \mbox{where} \\
& \quad y = 
\left\{\!\!
\begin{array}{l}
\argmax_k\; s\left(E^v(i^t_k),E^s(d^s_q)\right)\;\; \mbox{(image-based)} \\
\argmax_k\; s\left(E^t(d^t_k),E^s(d^s_q)\right)\;\; \mbox{(description-based)}
\end{array}
\right.
\end{split}
\end{equation}
Table \ref{table:knn} shows the results of the nearest-neighbor methods. "with dec." represents a multimodal space jointly trained with decoder while others indicate independently trained ones (\ie, the first step in the two-step approach). For reference, we also noted the performance when we randomly sampled a description in $\mathcal{T}^t$.
As expected, the three-way model generally outperformed the two-way model.
Interestingly, we observed that the performance was further improved when we directly retrieved descriptions on the target side.
This fact indicates that descriptions projected into the multimodal space still represent some useful information not apparent in the images. 

We hypothesize that jointly optimizing multimodal embedding loss and decoder loss (end-to-end model) may result in a better multimodal space because
decoder learning can be a good constraint in a multi-task learning framework.
As shown in the result, "with dec." models generally achieve better performance on IAPR-TC12, but not on Multi30K.
This is reasonable because independently trained multimodal space is poor on IAPR-TC12 but relatively good on Multi30K as the comparison with "TFIDF + CNN feature" baseline suggests. 

\begin{table}[tbp]
 \caption{\label{table:knn}BLEU (BLEU+1) scores of nearest-neighbor methods. {\small "with dec."} represents when jointly optimized with decoder in end-to-end training.}
 \begin{center}
  \begin{tabular}{|l||c|c|}
   \hline
                               & IAPR-TC12 & Multi30K \\
   \hline\hline
   \multicolumn{3}{|c|}{De $\rightarrow$ En} \\\hline
   Random             &  7.3 (9.0) & 0.9 (9.7) \\
   TFIDF + CNN feature & 17.8 (17.9) & 0.8 (9.7) \\
   2-way              & 16.5 (16.4) &  2.3 (11.0) \\
   3-way~(image)      & 17.3 (17.7) & 3.4 (12.3) \\
   3-way~(description)    & \textbf{21.1} (\textbf{20.9}) & \textbf{4.8} (\textbf{13.5}) \\\hline
   3-way, with dec.~(image)       & 18.3 (18.7) & 4.0 (12.2) \\
   3-way, with dec.~(description) & \textbf{22.5} (22.7) & \textbf{4.9} (\textbf{13.6}) \\\hline
   \multicolumn{3}{|c|}{En $\rightarrow$ De} \\\hline
   Random             &  3.8 (7.5) & 0.8 (9.9) \\
   TFIDF + CNN feature& 15.3 (15.4) & 1.6 (10.6) \\
   2-way              & 12.3 (13.0) & 2.1 (11.1) \\
   3-way~(image)      & 14.2 (15.5) & 3.1 (12.7) \\
   3-way~(description)    & \textbf{17.0} (\textbf{17.7})& \textbf{5.0} (\textbf{14.0}) \\\hline
   3-way, with dec.~(image)       & 14.6 (15.5) & 2.8 (12.5) \\
   3-way, with dec.~(description) & \textbf{19.4} (\textbf{20.2}) & \textbf{5.2} (\textbf{14.1}) \\\hline
  \end{tabular} 
 \end{center}
\end{table}

\vspace{-2mm}
\subsection{Main Results and Discussion}
\vspace{-1mm}
Table \ref{table:results} shows a detailed comparison of our approach in different configurations.
Comparing with the baselines (Table \ref{table:baseline}), our best results are roughly comparable to sequence-to-sequence models when the number of parallel sentences are limited to about 20\% as large as our monolingual ones.
We summarize our findings below.

A comparison of model topologies shows that the three-way models generally outperformed their two-way counterparts. However, when only images were feed-forwarded for training decoder, the differences in performance were subtle, and
the two-way model sometimes outperformed the three-way model.
The most attractive aspect of the three-way approach is that we can use both image and description for decoder training, which always provided the best results.
We can possibly utilize external monolingual corpora to further improve decoders, which we would like to investigate in our future work.

As for training strategy, end-to-end training generally achieved better results than the two-step approach, but the difference is not very large on Multi30K.
For IAPR-TC12, as the results of the nearest-neighbor experiment suggest, the multimodal space itself was relatively poor (sometimes outperformed by the TFIDF baseline).
In such a case, jointly optimizing the multimodal space (encoders) and the decoder seemed to significantly improve the performance. 
This result also corresponds to the observation in the previous section. 

\begin{table*}[btp]
 \caption{\label{table:results}BLEU (BLEU+1) scores comparison on different models and training strategies.}
 \begin{center}
  \begin{tabular}{|l|l|l||c|c|}
   \hline
   Topology & Training Strategy & Decoder training & IAPR-TC12 & Multi30K \\
   \hline\hline
   \multicolumn{5}{|c|}{De $\rightarrow$ En} \\\hline
   2-way & 2-step & image                & 23.6 (24.0) & 6.6 (17.0) \\
   2-way & end-to-end & image            & 23.2 (23.2) & 6.3 (17.0) \\
   3-way & 2-step & image                & 21.8 (21.7) & 7.3 (17.7) \\
   3-way & 2-step & description             & 21.3 (21.3) & 7.6 (18.4) \\
   3-way & 2-step & image + description     & 24.0 (25.6) & 8.0 (\textbf{18.9}) \\
   3-way & end-to-end & image            & 24.2 (24.3) & 7.6 (17.8) \\
   3-way & end-to-end & description         & 25.5 (26.2) & 8.1 (18.4) \\
   3-way & end-to-end & image + description & \textbf{26.0} (\textbf{26.7}) & \textbf{8.4} (\textbf{18.9}) \\\hline
   \multicolumn{5}{|c|}{En $\rightarrow$ De} \\\hline
   2-way & 2-step & image                & 19.0 (21.8) & 6.1 (17.2) \\
   2-way & end-to-end & image            & 19.6 (21.5) & 5.8 (17.8) \\
   3-way & 2-step & image                & 17.8 (20.0) & 7.5 (18.8) \\
   3-way & 2-step & description             & 18.5 (19.4) & 7.2 (18.8) \\
   3-way & 2-step & image + description     & 20.7 (22.8) & 7.6 (19.2) \\
   3-way & end-to-end & image            & 20.3 (21.9) & 6.8 (17.8) \\
   3-way & end-to-end & description         & 21.5 (22.7) & 7.5 (19.2) \\
   3-way & end-to-end & image + description & \textbf{23.0} (\textbf{24.3}) & \textbf{8.0} (\textbf{19.4}) \\
   \hline
  \end{tabular} 
 \end{center}
\end{table*}

\begin{figure*}
\begin{center}
\includegraphics[width=1.01\columnwidth]{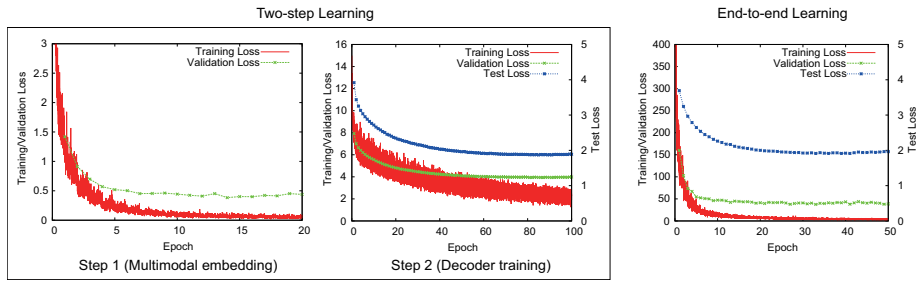}
\caption{Loss curves of English to German translation task on our three-way models (image + description) for the IAPR-TC12 dataset.} 
\label{fig:iaprtc12}
\end{center}
\end{figure*}

\begin{figure*}
\begin{center}
\includegraphics[width=1.01\columnwidth]{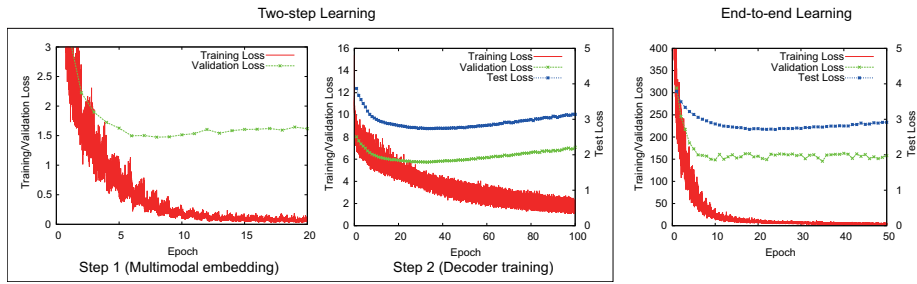}
\caption{Loss curves of English to German translation task on our three-way models (image + description) for the Multi30K dataset.} 
\label{fig:multi30k}
\end{center}
\end{figure*}

We show the loss curves of English to German translation task on our three-way models.
Figure \ref{fig:iaprtc12} and \ref{fig:multi30k} show the results on IAPR-TC12 and Multi30K, respectively.
Note that the training~(validation) loss cannot be directly compared to the test loss because they are based on entirely different criteria.
Nonetheless, we can see that validation loss and test loss converge in similar timings, making it possible to tune the network properly.
Another observation is that decoder training seems to be overfitting earlier on Multi30K.
This could be another reason that end-to-end approach showed no significant improvement on this dataset.

Finally, we demonstrate some qualitative results of zero-shot translation at Table \ref{table:examples}.
We observed that our method can actually translate many sentences correctly. 
Besides successful ones, we also see many interesting errors.
In many translations, although overall description of a scene is more or less relevant, attributes (\eg, color) and numbers of objects are often missed. This is reasonable because we are currently using only a single visual feature vector by global CNNs and therefore it is difficult to align fine-grained local information of images correctly. 
To tackle this problem, it would be promising to integrate more sophisticated object detection and segmentation methods in future. 
We also observe a number of small grammatical errors, possibly due to the lack of sufficient training data. We expect that this problem can be mitigated by utilizing external monolingual data in target language.

\begin{table*}[btp]
 \caption{\label{table:examples}Qualitative examples of German to English translation using our end-to-end three-way model (image + description). Ground truth English captions are noted in parentheses.}
 \begin{center}
  \begin{tabular}{|p{5.5cm}|p{5.5cm}|}
   \hline
   \textbf{Source (German)} & \textbf{Translation (English)} \\
   \hline\hline
   \multicolumn{2}{|c|}{\textbf{Successful translations}} \\\hline\hline
    ein radfahrer in einem gelben radtrikot, kurzer schwarzer radhose und mit einem graublauen helm, f\"{a}hrt auf einem gelben rennrad auf der linken stra{\ss}enseite eines highways; & a male cyclist wearing a yellow jersey, black cycling shorts and a grey and blue helmet, riding a yellow racing bike on the left hand side of a country \textbf{road}; \\
                            & (a male cyclist wearing a yellow jersey, black cycling shorts and a grey and blue helmet, riding a yellow racing bike on the left hand side of a country \textbf{highway};)\\\hline
   zuschauer sitzen auf einem braunen sandstrand im vordergrund; ein mann reitet auf einer brechenden welle im meer dahinter; ein grauer himmel im hintergrund; & \textbf{tourists} are sitting at a brown sandy beach in the foreground; a man is surfing a breaking wave in the sea behind it; a grey sky in the background; \\
                            & (\textbf{spectators} are sitting at a brown sandy beach in the foreground; a man is surfing a breaking wave in the sea behind it; a grey sky in the background;) \\\hline\hline
   \multicolumn{2}{|c|}{\textbf{Attribute, counting errors}} \\\hline\hline
   ein dunkelh\"{a}utiges m\"{a}dchen mit langen schwarzen haaren und einem blauen pullover steht an einem braunen ufer im vordergrund; ein dunkelblauer see dahinter; wei{\ss}e wolken an einem blauen himmel im hintergrund; & a dark-skinned \textcolor{red}{boy} with long black hair and a \textcolor{red}{white sweater} is standing in a brown shore in the foreground; a dark blue lake behind it; white clouds in a blue sky in the background; \\
          & (a dark-skinned \textcolor{blue}{girl} with long black hair and a \textcolor{blue}{blue pullover} is standing on a brown shore in the foreground; a dark blue lake behind it; white clouds in a blue sky in the background;)\\\hline
   eine frau in einem rosa kleid h\"{a}lt ein baby. & a \textcolor{red}{young} in a \textcolor{red}{blue shirt} is holding a baby.\\
                                                 & (a \textcolor{blue}{woman} in a \textcolor{blue}{pink skirt} is holding a baby.) \\\hline 
   drei m\"{a}nner stehen auf einem siegerpodium mit einer gelbblauwei{\ss}en wand dahinter; & \textcolor{red}{a men} are standing on a podium with a yellow, blue and white wall behind it; \\
                                                                                      & (\textcolor{blue}{three men} are standing on a podium with a yellow, blue and white wall behind it;) \\\hline
   ein blondes kind schaukelt auf einer schaukel. & a little boy is on a swing. \\
                                                   & (a \textcolor{blue}{blond} child \textcolor{blue}{swinging} on a swing.) \\\hline\hline
   \multicolumn{2}{|c|}{\textbf{Gramatical errors}} \\\hline\hline
    eine braune berglandschaft mit einigen schneebedeckten bergen; & a brown mountain landscape with a \textcolor{red}{snow snow} covered mountains; \\
                                                                    & (a brown mountain landscape with a few snow covered peaks;) \\\hline
    blick auf die h\"{a}user einer stadt am meer mit grauen wolken an einem blauen himmel im hintergrund; & view of \textcolor{red}{a} houses of a city at \textcolor{red}{a} sea; \textcolor{red}{a} clouds in the city sky in the background; \\
                                                                                             & (view of the houses of a city at the sea with grey clouds in a blue sky in the background;)\\\hline
  \end{tabular} 
 \end{center}
\end{table*}

\vspace{-1mm}
\section{Conclusion}
\vspace{-1mm}
In this work, we tackled a challenging task of training an NMT system from just monolingual data containing multimedia side information.
Unlike many previous studies that used multimedia simply in addition to texts as inputs to reinforce machine translation, 
we used no parallel corpora for training or image inputs in the testing phase. 
Our system was made possible by training multimodal encoders to share common modality-agnostic semantic representation using images as the pivot. 
We compared several possible implementations and showed the feasibility of our approach.
Notably, we found the three-way model to be particularly promising in terms of both performance and flexibility in handling various modality-specific data.
Although our target in this paper was a fully unsupervised setup, we can naturally include some parallel data in a semi-supervised manner 
or external monolingual text corpora in the target language to further enhance performance, which is an attractive direction for future research.

Of course, the experimental results also suggest that we have a long way to go. There is still a significant gap in performance as compared to supervised sequence-to-sequence baselines.
We expect this gap to further reduce as we use more expressive visual encoders, powerful attention mechanisms, and multimodal learning methods, 
all of which have remarkably improved in recent years. 
Moreover, our current method is intrinsically limited to the domain where texts can be grounded to visual content, which is not always the case in generic documents.
We would like to extend our approach to handle other side information and investigate how far we can go on automatically crawled noisy Web data, 
which is an important milestone to realizing true zero-resource MT utilizing abundant multimedia monolingual documents on the Web.

\section*{Acknowledgment}
This work is supported by JSPS KAKENHI Grant Number 16H05872 and JST CREST.
\bibliographystyle{spbasic}      
\bibliography{ImageRecognition,MachineLearning,DeepLearning,FGVC,NLP}

\end{document}